\documentclass{article}
\usepackage{iclr2026_conference}
\iclrfinalcopy

\usepackage{graphicx} 
\usepackage[hidelinks]{hyperref}

\usepackage{xcolor} 
\usepackage[toc,page]{appendix}
\usepackage{float}
\usepackage{algorithm}
\usepackage{adjustbox}
\usepackage{amssymb}
\usepackage{authblk}
\usepackage{algpseudocode}
\usepackage{amsmath}
\usepackage{booktabs}
\usepackage{subcaption}
\usepackage{hyperref}
\usepackage{tabularx}
\usepackage{natbib}

\title{Mnemosyne: An Unsupervised, Human-Inspired Long-Term Memory Architecture for Edge-Based LLMs}
\author{Aneesh Jonelagadda \quad Christina Hahn \quad Haoze Zheng \quad Salvatore Penachio \\
\small{Kaliber AI, Research Division} \\
\texttt{\small{\{a.jonelagadda, c.hahn, h.zheng, s.penachio\}@kaliber.ai}}
}

\date{September 2025}

\begin{document}

\maketitle

\begin{abstract}
 Long-term memory is essential for natural, realistic dialogue. However, current large language model (LLM) memory systems rely on either brute-force context expansion or static retrieval pipelines that fail on edge-constrained devices. We introduce Mnemosyne, an unsupervised, human-inspired long-term memory architecture designed for edge-based LLMs. Our approach uses graph-structured storage, modular substance and redundancy filters, memory committing and pruning mechanisms, and probabilistic recall with temporal decay and refresh processes modeled after human memory. Mnemosyne also introduces a concentrated ``core summary" efficiently derived from a fixed-length subset of the memory graph to capture the user's personality and other domain-specific long-term details such as, using healthcare application as an example, post-recovery ambitions and attitude towards care. Unlike existing retrieval-augmented methods, Mnemosyne is designed for use in longitudinal healthcare assistants, where repetitive and semantically similar but temporally distinct conversations are limited by naive retrieval. In experiments with longitudinal healthcare dialogues, Mnemosyne demonstrates the highest win rate of 65.8\% in blind human evaluations of realism and long-term memory capability compared to a baseline RAG win rate of 31.1\%. Mnemosyne also achieves current highest LoCoMo benchmark scores in temporal reasoning and single-hop retrieval compared to other same-backboned techniques. Further, the average overall score of 54.6\% was second highest across all methods, beating commonly used Mem0 and OpenAI baselines among others. This demonstrates that improved factual recall, enhanced temporal reasoning, and much more natural user-facing responses can be feasible with an edge-compatible and easily transferable unsupervised memory architecture.
\end{abstract}

\section{Introduction}

Memories are referred to as the “database of the self” \citep{Mahr_Csibra_2018,CONWAY2005594}. For a natural language exchange between two agents to be human-like, both parties must have a reliable sense of long-term memory. This is even more true for in-person edge AI agents, which must produce responses that are as natural, helpful, and accurate as possible during physical interactions with the user. 

However, LLMs are constrained by fixed context limits which cause earlier content to be excluded from its context once conversation length exceeds the limit. This causes predictable memory loss during an exchange between two agents, which is a functionality gap. Related works \citep{yan2025memoryr1enhancinglargelanguage, chhikara2025mem0buildingproductionreadyai, liu2025comprehensivesurveylongcontext} attempt to address this issue but have key weaknesses for our use case, such as not encoding temporal information, requiring a significant data and/or development overhead for applications in new domains, or not being viable for edge-constrained devices.

Our goal is to have a lightweight model that can refer to past events, either responding to a user with the correct context or bringing up the appropriate context when relevant. To this end, we propose \textit{Mnemosyne}, an unsupervised, graph-augmented storage and retrieval architecture for edge-based LLMs, integrating modular intake filters for substance, redundancy, and pruning. The system supports probabilistic recall, temporal decay modeled after human forgetting \citep{10.1371/journal.pone.0120644}, and boosting mechanisms for redundant memories, enabling it to behave more like a human memory system rather than a static retrieval pipeline. In addition, the architecture encapsulates core personality extraction as a deep memory module, and introduces novel query-relevance steering mechanisms such as naturalized time-delta placement keywords. Ultimately, Mnemosyne is a context engineering architecture, controlling what enters the downstream LLM’s context window. To our knowledge, this is the first memory system that combines modular intake filtering, dynamic graph-structured storage, and human-inspired recall dynamics for long-term LLM memory on constrained devices.

Mnemosyne is uniquely positioned to augment conversational agents in settings where on-premise edge agents interact with users regularly and collect interaction data. In the standard LoCoMo benchmark \citep{maharana2024evaluating}, Mnemosyne achieves state-of-the art performance in temporal reasoning by improving on alternative methods with a J-score of 60.42\%. Additionally, it achieves a much higher win rate in blind human evaluations of 65.8\% vs a naive RAG baseline's 31.1\%, demonstrating both effective factual recall and much more natural user-facing responses.
\section{Related Work}\label{sec:related work}

LLMs based on the Transformer architecture \citep{NIPS2017_3f5ee243} can now scale context windows into the hundreds of thousands of tokens \citep{xiao2024efficientstreaminglanguagemodels}. However, quadratic attention costs and KV-cache memory scaling make this approach unfeasible on edge devices, where memory and compute are limited. Long-context models, with context windows of 128k to over 1M tokens, can afford to retrieve and process orders of magnitude more memories within a single prompt than an Small Language Model (SLM) with a context window of $\sim$4k tokens. \citep{subramanian2025smalllanguagemodelsslms}. Thus, brute-force long-context methods used by Long-Context Language Models (LCLMs) remain impractical for on-device deployment. Episodic memory buffers maintain a sliding window of the most recent conversational turns, as seen in commercial chatbots like \href{https://pi.ai/onboarding}{Pi} and \href{https://replika.com/}{Replika}, but these methods discard older information regardless of salience.

Our design allows the system to maintain a long-term history with a size constrained by the total available system memory, which scales more gracefully than the quadratically-scaling limitations of the LLM's attention mechanism. By selectively retrieving and injecting only the most salient nodes into LLM's context mechanism, we bypass the bottleneck of a limited context window on resource-constrained devices.

RAG systems \citep{lewis2021retrievalaugmentedgenerationknowledgeintensivenlp} perform efficiently context-wise and are effective given a semantic variety of potential topics. Our use case, however, is a single-user domain-specific interaction agent that generates conversations with significantly less semantic diversity than those seen in common general-purpose vector stores. These interactions' importance wax and wane over time, necessitating a more dynamic approach. The method proposed in this paper is fundamentally context engineering-based, but is both more generalizable to memory as a whole and more specialized to the edge and constrained semantic domains.

\subsection{LLM Memory Systems}\label{sec: memrelatedwork}
More sophisticated techniques use structured external memory, such as MemGPT \citep{packer2024memgptllmsoperatingsystems} and \href{https://langchain-ai.github.io/langmem/}{LangMem}, which query a vector store. While this is a step in the right direction, traditional vector similarity search is inadequate for our domain due to the high semantic similarity of memories; very subtle but crucial differences between recurring events (e.g., daily wound care reports) are often lost in naive vector comparisons, leading to noisy retrieval. 

\begin{table}[ht]
\caption{Comparison of Mnemosyne with past related methods. $\sim$ indicates partial support.}
\centering
\begin{adjustbox}{width=1\textwidth}
\label{tab:relatedworks}
\small
\begin{tabular}{lcccccc}
\hline
\textbf{System} & \textbf{Structured} & \textbf{Semantic} & \textbf{Temporal} & \textbf{Edge-} & \textbf{Redundancy} & \textbf{Forgetting} \\
                & \textbf{Memory}     & \textbf{Recall}   & \textbf{Dynamics} & \textbf{Feasible} & \textbf{Handling} & \textbf{Model} \\
\hline
Long-Context    & X         & \checkmark & X         & X         & X        & X \\
Naive RAG       & \checkmark & \checkmark & X         & \checkmark         & X        & X \\
MemGPT/LangMem  & \checkmark & \checkmark & X         & X         & $\sim$   & X \\
Memory-R1       & \checkmark & \checkmark & X         & $\sim$         & \checkmark & \checkmark \\
Mem0            & \checkmark & \checkmark & $\sim$         & $\sim$         & \checkmark   & $\sim$ \\
\textbf{Mnemosyne} & \checkmark & \checkmark & \checkmark & \checkmark & \checkmark & \checkmark \\
\hline
\end{tabular}
\end{adjustbox}
\end{table}

As shown in Table \ref{tab:relatedworks}, prior systems address fragments of the long-term memory problem. Refinements to long-term memory, such as those proposed in Memory-R1 \citep{yan2025memoryr1enhancinglargelanguage} and Mem0 \citep{chhikara2025mem0buildingproductionreadyai}, inspire aspects of Mnemosyne's design but still have slight functionality gaps. For example, Mem0 uses a rules-based cache management strategy as a forgetting mechanism, updating previous information if it is contradicted by new information. The system proposed in this paper, however, incorporates temporal decay and refresh/rewind mechanisms modeled after human memory studies \citep{Mahr_Csibra_2018, CONWAY2005594}. Thus, our forgetting process is probabilistic, continuous, and tunable. Memory-R1 introduces a comprehensive forgetting mechanism similar to the one we propose, but lacks any kind of temporal encoding in its stored memories. Additionally, like MemGPT/LangMem, Memory-R1 was not built to be edge-compliant; despite potential for refactoring, its edge efficacy has not been tested to our knowledge. Memory-R1's further reliance on reinforcement learning for add/delete policy optimization makes it less interpretable and requires specialized training runs with carefully crafted reward signals. This makes it expensive to adapt and deploy to new domains. In contrast, Mnemosyne is unsupervised, which makes it lightweight and domain-transferable with minimal overhead.

\section{Methods}

Our memory system is designed for high performance, storing all necessary graph information on an in-memory database for its speed advantages (see Section \ref{sec: implementation}). At a high level, our system is comprised of (1) commitment, (2) recall, (3) asynchronous updates to the system's core supersummary, and (4) a pruning module. 

\begin{figure*}[ht]
    \centering
    \begin{subfigure}[b]{0.5\textwidth}
        \centering
        \includegraphics[width=1.15\textwidth]{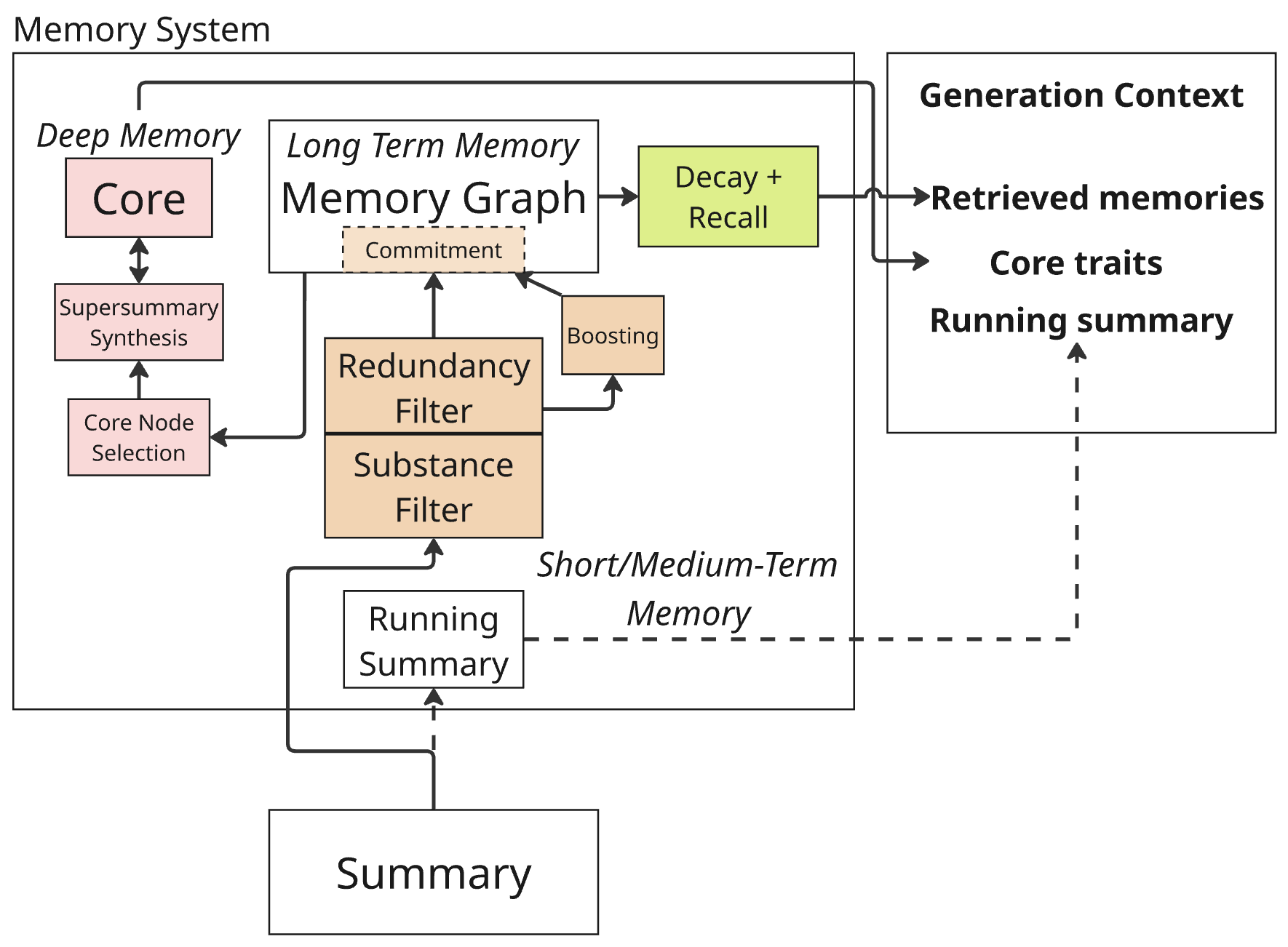}
        \label{fig:high-level-structure}
        \caption{}
    \end{subfigure}%
    ~ 
    \begin{subfigure}[b]{0.5\textwidth}
        \centering
        \includegraphics[width=0.7\textwidth]{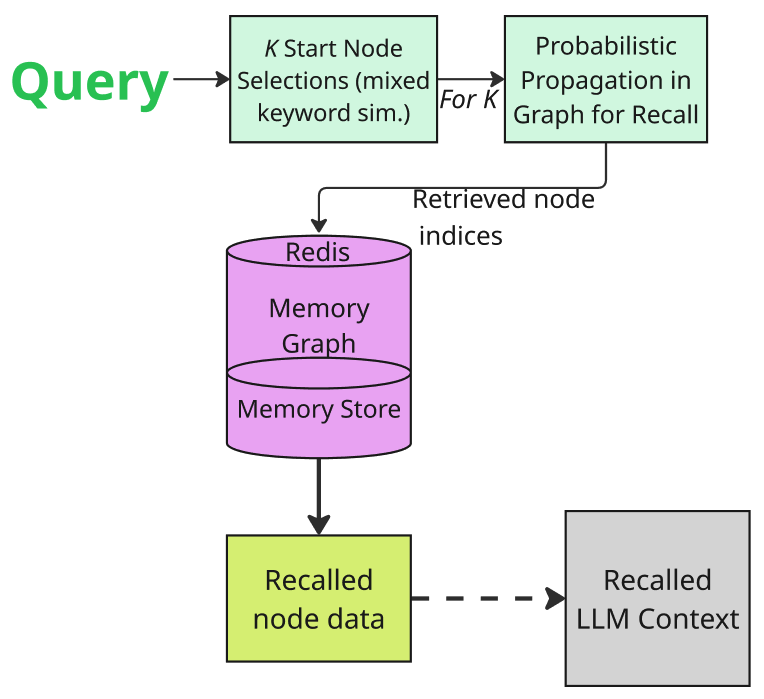}
        \caption{}
    \end{subfigure}%
    \\
    \begin{subfigure}[b]{\textwidth}
        \centering
        \includegraphics[width=\textwidth]{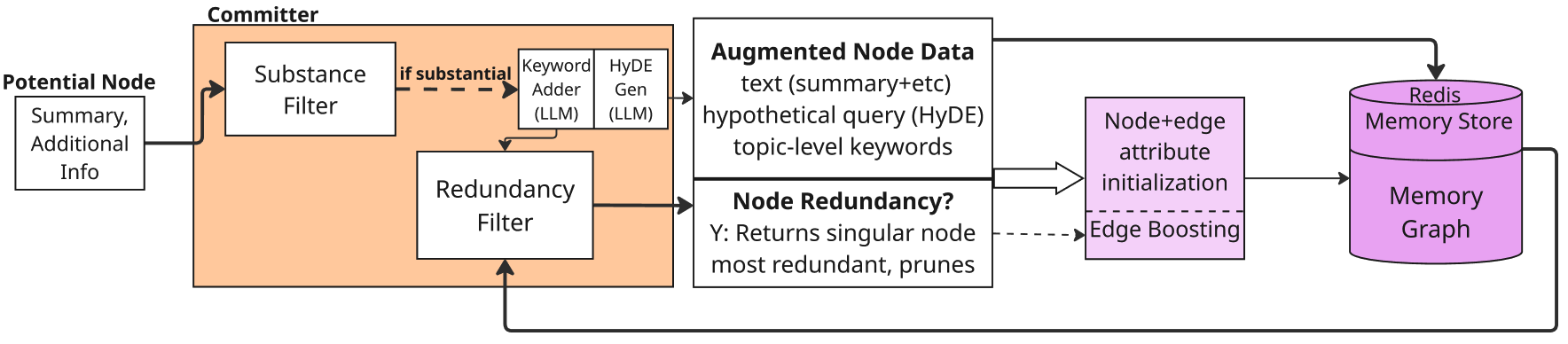}
        \caption{}
    \end{subfigure}
    
    \caption{High-level layout of Mnemosyne: (a) the entire memory unit, (b) the recall process, and (c) the commitment process.}
    \label{fig:high-level-layout}
\end{figure*}

\begin{figure*}[ht]
    \centering
        \includegraphics[width=0.85\textwidth]{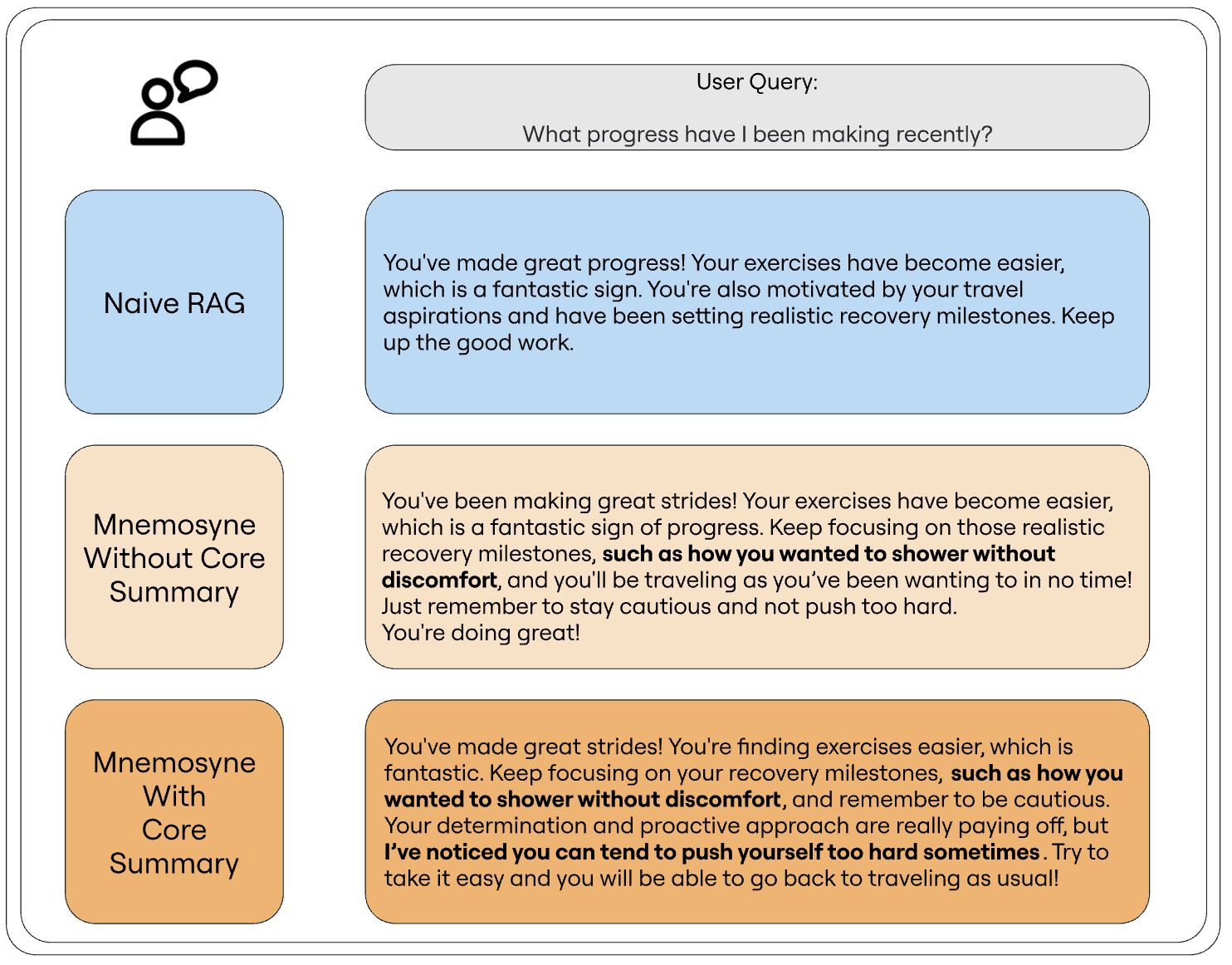} 
        \caption{Example queries and responses for Mnemosyne with and without the core summary module compared to a RAG baseline. Key differences in responses are highlighted in bold, demonstrated long-term memory behavior and core user understanding.}
        \label{fig: queryanswer}
\end{figure*}

The memory system inputs are interaction summaries. These are generally atomic to the interaction, but we also recommend ingestion of a shorter-term iteratively updated running summary. The summaries are then asynchronously passed into the commitment algorithm, which either discards or commits the input into the memory graph as a node by using the core summary as a salience baseline. Upon a user query, the recall algorithm selects a start node and probabilistically traverses the graph, selecting a set of the most relevant nodes and converting it into formatted LLM generation context. Finally, the core summary is periodically updated to ensure that newer conversations are accounted for. This process is also asynchronous. The high level layout of Mnemosyne can be seen in Figure \ref{fig:high-level-layout}. Example queries and responses of Mnemosyne compared to a RAG baseline are shown in Figure \ref{fig: queryanswer}.

\subsection{Commitment}\label{sec: commitment}

Our memory system's commitment algorithm contains two filters: (1) an LLM-based substance filter that discards unimportant information, and (2) a traditional ML-based redundancy filter that marks and prunes redundant conversations. From there, the Algorithm \ref{alg:main_commit} is dedicated to storing all information that will be necessary during recall to ensure high performance. 
\begin{algorithm}[]
\caption{Commitment Algorithm}\label{alg:main_commit}
\begin{algorithmic}[1]
\Procedure{CommitMemory}{$s_{new}, d_{new}, G$}
    \If{$\neg \Call{HasSubstance}{s_{new}, G}$}
        \State \textbf{return} $G$ \Comment{Discard summary, no changes made}
    \EndIf
    
    \State $n_{new} \gets \Call{InstantiateNode}{s_{new}, d_{new}}$
    \State $n_{red} \gets \Call{FindMostRedundantNode}{n_{new}, G}$ \Comment{Finds most similar node}
    
    \If{$n_{red}$ is \textbf{null}} \Comment{Case 1: No redundant node found}
        \State $G.\Call{AddNode}{n_{new}}$
    \ElsIf{$n_{red}.\text{is\_paired}$ is \textbf{null}} \Comment{Case 2: Redundant with an unpaired node}
        \State $G.\Call{AddNode}{n_{new}}$
        \State $G.\Call{SetPair}{n_{new}, n_{red}}$
        \State $e_{n_{red}n_{new}}.\text{boost}\gets\Delta_e(t)$
    \Else \Comment{Case 3: Redundant with an already paired node}
        \State $n_{pair} \gets G.\Call{GetNode}{n_{red}.\text{is\_paired}}$
        \State $n_{to\_keep} \gets \Call{SelectOldest}{n_{red}, n_{pair}}$
        \State $n_{to\_remove} \gets \Call{SelectNewest}{n_{red}, n_{pair}}$
        \State $G.\Call{RemoveNode}{n_{to\_remove}}$
        \State $G.\Call{AddNode}{n_{new}}$
        \State $G.\Call{SetPair}{n_{new}, n_{to\_keep}}$
        \State $e_{n_{red}n_{new}}.\text{boost}\gets\Delta_e(t)$
    \EndIf
    
    \State $\Call{ConnectNewNode}{n_{new}, G}$ \Comment{Create edges to similar nodes}
    \State \textbf{return} $G$
\EndProcedure
\end{algorithmic}
\end{algorithm}

Because not every memory has substance, our commitment algorithm has an LLM-based filtering mechanism to discard irrelevant incoming conversation summaries. Using the conversation summary, any other ancillary summaries, and the core summary as inputs, an LLM judges if the contents of the conversation are substantial or not. We can thus formulate the substance filter as a binary classifier. We define ``substantial" conversations in the instructions to the LLM as conversations that include information from one of the following categories: (1) the patient's medical condition, treatment, or care, (2) the patient's clinical status, behavior, or needs, and (3) details about the patient's personality. Note that routine or mundane conversations are considered non-substantial. If interactions are deemed substantial, they get forwarded to the rest of the committer algorithm.

\subsubsection{Redundancy Filter and Rewind Calculation} \label{sec:redundancy}

Even if conversations are deemed substantial, many similar conversations in the same memory graph may bloat the system and decrease performance. Thus, the commitment process has a redundancy filter to detect redundant incoming nodes and pair them with an existing node. Taking inspiration from the primacy-recency effect \citep{mayo1964cognitive, morrison2014primacy}, we discard the more recent of the two existing nodes if an incoming conversation is deemed redundant to a node already paired.

We calculate redundancy via a combination of: (1) the mutual information \citep{duncan1970calculation} between the two nodes' conversation summary embeddings and (2) the Jaccard similarity between the keywords of those conversation summaries. Implementation-wise, entropy and joint entropy for textual documents can be calculated programmatically by binning fixed-length text embeddings for the mutual information calculation.

Let node $n$ be the incoming node, $m$ be some existing node in the memory graph $G$. Because we are operating on the embeddings of conversation summaries $n_s, m_s\in\mathcal{S}$, let the embedding function $f$ be defined as $f(n) =\mathbf{e_n}$. Given some constant weight $\alpha_{\text{NMI}}\in [0,1]$ and the redundancy score calculation $\text{RS}(n, m)$, we calculate $m^+$, the existing node with the maximum redundancy score:
\begin{gather*}
    \text{RS}(n, m) = \alpha_{\text{NMI}} \cdot \text{MI}( \mathbf{e_{n_s}}, \mathbf{e_{m_s}}) + (1-\alpha_{\text{NMI}})\cdot \text{JS}(\mathbf{e_{n_k}}, \mathbf{e_{m_k}})\\
    m^+ = \text{arg}\max_{m \in G}\text{RS}(n,m)
\end{gather*}

 We finally define some minimum redundancy score threshold \text{RS}\textsubscript{\text{min}} to set the redundant node $m^+$. If $m^+$ is already paired, $m^+$ is set to the older node in the pairing and the newer one is discarded.

Repeated memories are ``refreshed" and retained for longer periods of time \citep{ebbinghaus1913memory}. Thus, when a memory has been determined to be redundant by the redundancy filter, we introduce a ``boosting" or rewind mechanism that reverses the decrease in selection probability caused by temporal decay during recall.

This rewind value is calculated during commitment and stored for use during recall. We use a sigmoidal function which plateaus at higher timescales to bound the boosting mechanism and ensures some maximal boost which is within the domain of the decay function.
\begin{equation}\label{eq: rewind}
    \Delta_e(t = t_\text{curr})=\Delta_\text{max}\left(\displaystyle\frac{1}{1+\exp(-t+e_\text{boosted}+t_\text{crit})}\right)
\end{equation} where $e=e_{nm}$ be the edge being boosted, $t$ the current time $t_\text{curr}$, and $e_\text{boosted}$ the time of the last edge boost. Given constants $t_\text{crit}$, which determines how long the function takes to plateau, and $\Delta_\text{max}$, which is the maximum rewind value. Further, the nonlinear behavior near zero de-incentivizes frequent updates to the same memory, which is important for encouraging diversity among the selected memories between user queries. Since small $t$ in this case corresponds to a short time since the most recent boost, this behavior resembles biological brains' desensitization to frequent stimulus \citep{thompson10model, groves1970habituation}.

\subsubsection{Graph Construction and Decay Calculation}\label{sec: graph construction}

We construct node and edge instances only with lightweight information necessary for critical graph operations (see Table \ref{tab:graph_attr}). Hypothetical queries, taking inspiration from HyDE \citep{gao2023precise}, are also generated and embedded during commitment to close the structural difference between the user's query and the graph content summaries during recall. Exact implementation details can be found in Section \ref{sec: implementation} and Appendix.

Once the input has been judged substantial by the substance filter, we create the node instance $n$ and create edges between $n$ and the existing nodes in the memory graph $G$ based on similarity. Our similarity score function is a weighted sum of (1) the cosine similarity between the two nodes' conversation summary embeddings, and (2) the Jaccard similarity between the keywords of those conversation summaries. 

Given a keyword mixing weight $\alpha_{\text{key}} \in [0,1]$ similar to $\alpha_{\text{NMI}}$ in Section~\ref{sec:redundancy}, our similarity score function is
\[
\operatorname{sim}(n, m) 
= \alpha_{\text{key}}
\cdot\operatorname{CS}(\mathbf{e_{n_s}}, \mathbf{e_{m_s}}) + (1-\alpha_{\text{key}})\cdot \operatorname{JS}(\mathbf{e_{n_s}},\mathbf{e_{m_s}})
\]
After introducing some minimum edge creation threshold \text{edge}\textsubscript{min}, we can then define $e_{nm}$ as the edge between the new node $n$ and some existing node $m\in G$ with weight given by $\operatorname{sim}(n, m)$.

Another intended functionality of our memory system is to emulate human memory by ``forgetting" older conversations at a natural rate. The majority of the legwork for this forgetting mechanism is done by ``decaying" the selection probability of older memories during recall. We account for refreshed memories by fetching the rewind value $\Delta_m(t)$ stored during commitment. Given some edge $e$ and current time $t=t_\text{curr}$, $e$'s effective age $e_\text{eff}$ can be defined as
$e_\text{eff} = t - e_\text{t\_init} - \Delta_e(t)$. From here, given $a$, the middle of the sigmoid, $b$, the sigmoid's ``steepness" (lower values indicate steeper dropoff), $c$, the transition point of the linear correction domain, and $d$, the floor decay value, our decay function can be formalized as

\begin{equation}\label{eq: decay}
    \tau(e_\text{eff}) = \begin{cases}
    \frac{1\ -\ d}{1\ +\ e^{\frac{\left(e_\text{eff}-a\right)}{b}}} &  c \leq e_\text{eff} \\[4ex]
    -\frac{e_\text{eff}\left(1-\tau\left(c\right)\right)}{c} &  0\leq e_\text{eff} \le c
    \end{cases}
\end{equation}
since we intend our function to plateau above 0 at higher timescales and ensure that past conversations are never fully forgotten. To ensure consistency in retrieval, we additionally enforce the edge weight's value to be 1 at $t=0$ regardless of parameters by imposing a linear correction to the modified reverse sigmoid function's early values. 

\begin{figure}\label{fig:rewind}
    \centering
    \begin{minipage}{0.45\textwidth}
        \centering
        \includegraphics[width=0.9\textwidth]{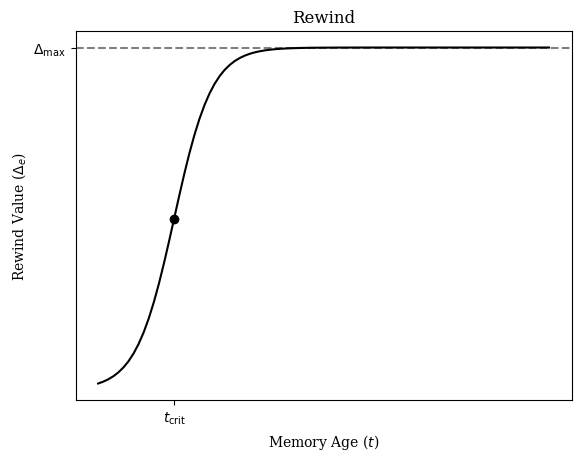} 
        \caption{Graph of the rewind function.}
    \end{minipage}\hfill
    \begin{minipage}{0.45\textwidth}
        \centering
        \includegraphics[width=0.9\textwidth]{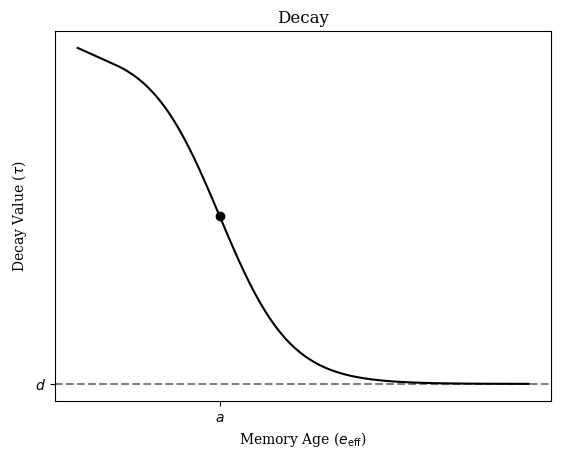} 
        \caption{Graph of the decay function.}
    \end{minipage}
\end{figure}

\subsection{Recall}\label{sec: recall}

The recall algorithm is responsible for retrieving the set of conversation summaries that will be returned as LLM context, accomplished via probabilistic traversal of the memory graph with modified edge weights. Our algorithm is inspired by neuron activation propagation, where neurons fire action potentials and influence nearby neurons \citep{feinerman2005signal}. The signal decays as it travels through the nervous system \citep{feinerman2005signal, kharas2022brain}, which we emulate through our algorithm's passive decay in selection probability past the start node.

We begin by selecting the node $m^*$ most similar to the user query. The hypothetical query generated during commitment is used here for the comparison to eliminate the difference in answer space between the user query and the graph content. We use a weighted sum of two similarity scores $S_\text{meta}$ and $S_\text{query}$ to select the start node. $S_\text{meta}$ is calculated with the embedded user query $u_q$ and a concatenation of (1) the naturalized time delta (e.g., \texttt{"last week"}) $m_{\Delta}$ between the current time $t$ and the initialization time $t_\text{init}$ of node $m$, (2) a natural language description of domain-specific state(e.g.,\texttt{"post-surgery recovery phase"}) and (3) relevant keywords $m_k$. $S_\text{query}$ is calculated with the embeddings of user query $u_q$, the hypothetical query $m_q$, and the summary text $m_s$. $S_\text{meta}$ and $S_\text{query}$ can be defined as
\begin{gather*}
S_\text{meta}=\text{CS}\left(\mathbf{e_{u_q}}, \mathbf{e_{m_\Delta + m_k}}\right) \quad S_\text{query}=\text{CS}\left(\mathbf{e_{u_q}}, \mathbf{e_{m_q + m_s}}\right) \\
S(m)=\alpha_{\text{meta}} \cdot S_\text{meta} + (1-\alpha_{\text{meta}})\cdot S_\text{query} \\
m^* = \text{arg}\max_{m \in G}S(n,m)
\end{gather*}
We would like to highlight the novelty of incorporating temporal language into the similarity computation rather than treating time as a raw numeric feature. We use the naturalized time delta because as time goes on, the user will refer to the memory contained in $m$ with relative temporal qualifiers instead of the present tense. Thus, it is necessary to account for during the recall process in order to accurately retrieve temporally situated memories. 

After selecting the start node $m^*$, we proceed with the recall algorithm by traversing the graph DFS-style via edge weights (see Section \ref{sec: graph construction}). The selection probability of the next node $m$ is modified by temporal decay and rewind mechanisms, and if selected, $m$ is added to a set of final nodes $R$ that will be returned, with their contents formatted as LLM context. Let $n$ be the current selected node during traversal, $m$ the next neighbor of $n$, and $e_{nm}$ the weight of the edge connecting them. Given an arbitrary exploration hyperparameter $\mu$, the selection probability of node $m$ can be defined as
\[
  P(m)= \mu \cdot e_{nm}\cdot\tau(e_\text{eff})
\]
We begin by calculating $P(m)$ for each neighbor $m\in N(m^*)$ and generating a random number $r\in[0, 1]$. If $r< P(m)$, $m$ is added to $R$ and $N(m)$ are recursively explored. The overall traversal algorithm is shown in Algorithm \ref{alg: traversal}.
\begin{algorithm}
\caption{Probabilistic Traversal}\label{alg: traversal}
\begin{algorithmic}[1]
\Statex \textbf{Input:} Query $q$, Memory graph $G$, Exploration hyperparameter $\mu$
\Statex \textbf{Output:} List of recalled node indices $R$

\Statex
\Procedure{RecallTraversal}{$n_{curr}, G, R, V, \mu$}
    \State $V \gets V \cup \{n_{curr}\}$
    \State $R.\Call{append}{n_{curr}.\text{index}}$
    \For{each neighbor $n_{k}$ of $n_{curr}$}
        \If{$n_{k} \notin V$}
            \State $e_{curr,k} \gets G.\Call{GetEdge}{n_{curr}, n_{k}}$
            \State $t_{eff} \gets \Call{CalculateEffectiveAge}{e_{curr,k}}$
            \State $\tau \gets \Call{DecayFunction}{t_{eff}}$ \Comment{Apply reverse sigmoid decay}
            \State $p_{choose} \gets e_{curr,k}.\text{weight} \times \tau \times \mu$
            \State $r \gets \Call{Random}{0, 1}$
            \If{$r < p_{choose}$}
                \State $\Call{RecallTraversal}{n_{k}, G, R, V, \mu}$
            \EndIf
        \EndIf
    \EndFor
\EndProcedure
\Statex
\end{algorithmic}
\end{algorithm}
We also separately introduce a pruning module which acts as the systems ``forgetter"; nodes are scored by a version of the selection probability without exploration term, and nodes below a pruning threshold are removed (see Equation \ref{eq:pruning score}). The pruning mechanism is only intended for use when the graph size approaches memory limits.

\subsection{Core Summary}\label{sec: core}

People possess personality quirks and remember emotionally resonant or distinctive events \citep{schmidt2006emotion}. These personalized details have been shown to help LLMs generate more natural and tailored responses to user queries \citep{qiu2025measuring, liu2025improving}. To represent these facts, we store a supersummary of past conversations that represent what the AI assistant has learned about the patient. We call this supersummary the ``core summary" and it is always injected into final LLM system prompt along with the recalled context. Most importantly, this core summary is generated on a fixed-length subset of the graph's memories. This allows for efficient and feasible longitudinal scaling; otherwise, if done naively, the supersummary-generating LLM would have to potentially ingest hundreds to thousands of memories at once, disallowing low-cost edge use of this system. 

The stored core summary is used in the substance filter as a baseline of comparison for ``what is substantial," and captures key details about the user. As stated in Section \ref{sec: commitment}, important information includes (1) the patient's medical condition, treatment, or care, (2) the patient's clinical status, behavior, or needs, and (3) details about the patient's personality. Our core summary algorithm begins by selecting a central subset of nodes $C$ from the memory graph $G$ subject to $|C| \leq $ a maximum node number $M$. In order to ensure appropriate span of topics represented in $C$, we first partition $G$ using k-means clustering \citep{lloyd1982least} on the node embeddings $\mathbf{e_n} \in \mathbb{R}^{d_{\text{embed}}}$, yielding $k$ cluster assignments.

All nodes across all clusters are then scored using a linear combination of (1) connectivity score $s_{\text{conn}}$, (2) boost score $s_{\text{boost}}$, (3) temporal recency score $s_{\text{rece}}$, and (4) information density score $s_{\text{ent}}$ to yield a hybrid score $s_\text{hybrid}(n;\theta)$ of node $n$. The coefficients $\theta$ in the linear combination are subject to the hyperparameter constraint $\sum_{i=1}^4\theta_i = 1$. Assuming each node $n$ has attributes embeddings vector $\mathbf{e_n}$, neighbors $N(n)$, initialization time $t_{\text{init}}$, and set of edges with boosts $\Delta_{n,m}$ $\forall m \in N(n)$,
\begin{align*}
    &s_{\text{conn}}(n)=\frac{|N(n)|}{|G|-1}&&s_{\text{boost}}(n)= \frac{|\{m \in N(n) : \Delta_{n,m} > 0\}|}{\max_{n' \in G} |\{m \in N(n') : \Delta_{n',m} > 0\}|}\\
    &s_{\text{rece}}(n) = \exp\left(-\frac{t_{\text{curr}}-t_{\text{init}}}{\lambda}\right)&&s_{\text{ent}}(n) = \frac{1}{1+H(\mathbf{e_n})}
\end{align*}

\noindent\textbf{Connectivity Score:}
We define the connectivity score $s_{\text{conn}}$  as the connectivity of the node $n$ normalized by the maximum possible connectivity of the graph $G$. This selects for highly central memories which semantically relate to other memories. 

\noindent\textbf{Boosting Score:}
In order to reward memory boosting due to highly related interactions being ingested (and deemed redundant) by our memory system, we count the number of boosted edges for node $n$ and normalize by the maximum number of boosted edges across all nodes in the current graph to form boost score $s_{\text{boost}}$.

\noindent\textbf{Recency Score:}
The temporal recency score $s_{\text{rece}}$ is introduced prioritize recent memories over old ones. We use a separate timescale than our decay function from Equation \ref{eq: decay} to ensure decoupling between the core memory system, which can have various scales of synchronicity and can be used independently for other use-cases such as personality extraction, and the regular long-term-memory graph-based recall with more frequent update synchronicity. We use a memory half-life $\lambda$ = 28 days. 

\noindent\textbf{Connectivity Score:}
Certain memories can be highly connected due to semantic generality, be recent, frequently related to incoming interaction data, but also contain little intrinsic information. A white noise process $W(n)$ thus propagates through our core system favorably. In order to counteract this, we introduce an information-density-based score $s_{\text{ent}}$ using the binned entropy $H(\mathbf{e_n})$.

Our design must allow for sufficient diversity across the core nodes, so we use the cluster assignments to select and aggregate the top scoring node per cluster yielding an initial selected set of core memories $C'$. Solely using one node per cluster in the core node subset can fully obfuscate the underlying distribution of memories, so we augment the subset with the top $k_2$ scoring nodes $C_\text{crit} \in G\setminus C'$ from the set of all nodes that have not been selected already. This yields the final core summary subset of nodes $C = C' \cup C_\text{crit} \text{ with } |C| = k+k_2$. Finally, if $|C| >$ the maximum node number $M$, we truncate the array simply by keeping the top $M$ most recent nodes.

For the updating process, the core summary $c$ is initialized as \texttt{"N/A"} when there are no available past conversations with the user. Periodically, the central subset $C$ of the graph is selected asynchronously, and raw conversation summaries corresponding to the nodes in $C$ will be passed into an LLM with the current core summary. There are two different instruction prompts: (1) initializes the core summary and only accounts for $s_c$, and (2) updates the core summary and accounts for both $c$ and $s_c$. Both prompts instruct the LLM to capture overall characteristics.

\section{Experimental Setup}\label{sec: implementation}

All experiments were conducted on machine equipped with a 12th Gen Intel Core i7-12700E processor (12 cores, 20 threads, up to 4.8GHz boost frequency), NVIDIA RTX A4000 GPU with 16GB VRAM, and 64GB DDR5 system memory.
Our hyperparameter configuration is provided in Table \ref{tab:hyperparameters}. All of our human evaluation, commitment and core operations involving LLMs were done with mistral-7b-instruct-v0.2.Q5\_K\_M \citep{jiang2023mistral7b}, our vector embeddings were generated by PubMedBERT \citep{Gu_2021,neuml_pubmedbert_embeddings}, and our external node attributes (see Section \ref{tab:node_attr}) and graph structure were stored on a Redis database \citep{redis}. Llama3.1-8B-Instruct \citep{grattafiori2024llama3herdmodels} was used as the generation LLM for LoCoMo benchmarking. 

\subsection{Memory System Benchmarking using LoCoMo}\label{sec: benchmarking}

We use the LoCoMo benchmark score \citep{maharana2024evaluating} to evaluate Mnemosyne's ability to recall both correct and contextually relevant information compared to other leading techniques listed in Section \ref{sec: memrelatedwork}. The LoCoMo benchmark provides a set of 10 scenarios, each containing a series of conversations and QA questions for benchmarking. The number of conversations in each series can range from 19 to 32 and each conversation has several turns. The query provided by the QA data (excluding the adversarial category) was thus used to perform recall (Section \ref{sec: recall}), feeding an LLM with context to generate an answer. We report the LLM-as-a-Judge (J) score using GPT-4o-mini \citep{hurst2024gpt} as the judge with a standard LLM-as-a-Judge prompt \citep{packer2024memgptllmsoperatingsystems} to compare with gold standard. This better captures semantic fidelity compared to the F1-score and BLEU-score, as noted in \citep{chhikara2025mem0buildingproductionreadyai}. Baseline metrics are cited directly from \cite{yan2025memoryr1enhancinglargelanguage}'s LLama3.1-8B-Instruct backbone results. The authors do not specify the judge model, though given similar baselines such as Mem0 it is likely GPT-4o-mini.

Our model was initially designed to tackle the difficulties of a high-density semantic space of specialized healthcare dialogues. In this domain of healthcare dialogues, conversations frequently revisit a narrow set of topics (e.g., pain, medication, recovery), creating a high degree of semantic overlap and redundancy. Our system's core mechanisms, such as redundancy detection using mutual information and the temporal boost/decay are specifically built to manage this overlap by consolidating recurring concepts and preserving subtle but important differences. The domain-agnostic utility of our architecture is demonstrated by the fact that the semantic space constraint is actually loosened when generalizing to a bigger domain such as the LoCoMo benchmark, where themes are more different and dispersed (e.g., travel, hobbies, family).

\subsection{Human Evaluation}

Retrieval metrics give us a quantifiable measure of accuracy, but they do not necessarily capture the ``naturalness" of LLM responses. To ensure a holistic evaluation, we also conducted a blind human study comparing three systems: (1) Mnemosyne with core summary, (2) Mnemosyne without core summary, and (3) Naive RAG baseline. 8 human annotators, including 1 non-response, were presented with 40 queries (e.g., ``What does the patient consider their most important personal value?") and three LLM responses to those queries corresponding to the three systems. The system order was randomly scrambled for each query. The annotators made pairwise judgments on their preferred response, with ties permitted when both responses were equally strong or weak. Using these pairwise judgments, we calculated the win rate (see Equation \ref{eq:win rate}) for each of the systems with a Wilson 95\% CI.

\section{Results}

\subsection{Retrieval Benchmarking}

Our benchmarking results are shown in Table \ref{tab:benchmarking}. Mnemosyne yielded the highest J-score among all baseline methods for the LoCoMo temporal reasoning and single-hop scores, with a second-highest overall score of 54.55\%. While OpenAI memory performed better for the single-hop category, the backbone LLM used is different and not edge-viable, so we discount this from the direct comparison. This indicates that Mnemosyne excels in temporal reasoning, which makes sense given its elaborate treatment of temporal dynamics, and its overall performance still beats most other baseline memory methods.

\begin{table}[ht]
\caption{LoCoMo Benchmark (J-score) of Mnemosyne compared to past related methods. Systems marked with * use Llama3.1-8B-Instruct as its generation LLM backbone, while † marks results reported directly in Yan et al. (2025).}
\centering
\begin{adjustbox}{width=1\textwidth}
\label{tab: locomo}
\small
\begin{tabular}{lcccccc}
\hline
\textbf{System} & \textbf{Single-Hop (\%)} & \textbf{Multi-Hop (\%)} & \textbf{Open-Domain (\%)} & \textbf{Temporal Reasoning (\%)} & \textbf{Overall (\%)} \\

\hline
Mem0\textsuperscript{*†}    & 43.93     & 37.35     & 52.27       & 31.4         & 45.68 \\
Zep\textsuperscript{*†}    &52.38      & 33.33      & 45.36      & 27.58         & 42.80  \\
MemGPT/LangMem\textsuperscript{*†}  & 47.26 & 39.81 & 48.38         & 30.94      & 44.18  \\
OpenAI       & \textbf{63.79} & 42.92 & 62.29         & 21.71         & 52.90 \\
Memory-R1\textsuperscript{*†}       & 59.83 & \textbf{53.01} & \textbf{68.78}         & 51.55        & \textbf{62.74} \\
Mnemosyne\textsuperscript{*}       & \textbf{62.78} & 49.53 & 60.42   & \textbf{53.03}  & 54.55 \\
\hline
\end{tabular}
\end{adjustbox}
\end{table}
\subsection{Human Evaluation}\label{sec:human eval}
Mnemosyne with the core summary outperforms both the ablated version of Mnemosyne without the core summary and the naive RAG baseline with a win rate of 65.8\%. Judges consistently preferred responses containing long-horizon and recurring details about the patient, which indicates that there is a notable advantage in maintaining a long-term persona-level supersummary in the memory system. Additionally, Mnemosyne without the core summary also outperformed the baseline. This suggests that the commitment gates, graph-structured recall, and human-inspired temporal mechanisms can improve response quality even without a deep summary module. However, the relative gap between the two Mnemosyne variants shows that core summarization is still necessary because individual memories alone seem to be less effective at reconstructing higher-order patterns.

An extension of this result is that perfect single- and multi-hop retrieval for long time-scales is not necessary to achieve perceived naturalness in an AI agent; a typical human will not arbitrarily remember what happened at a precise time several months ago. These findings validate our hypothesis that memory systems should be able to provide both specific and general information about the user for more natural and personalized responses from LLMs.

\begin{table}[ht]
\caption{Human evaluation results. The strongest result is bolded.}
    \centering
    \begin{tabular}{c|c|c}
         System & Win Rate (\%) & 95\% CI (Wilson)\\
         \hline
         \textbf{Mnemosyne w/ Core Summary} & \textbf{65.80} & [61.78, 69.61]\\
         Mnemosyne w/o Core Summary & 53.13 & [48.98, 57.22]\\
         Naive RAG & 31.07 & [27.38, 35.02]\\
    \end{tabular}
    \label{tab:qualitative}
\end{table}

\subsection{LoCoMo Benchmark Limitations}

We noticed a few worth-mentioning limitations behind the LoCoMo benchmark \citep{maharana2024evaluating} in terms of the gold standard answers’ quality and accuracy. Upon examining our model performance on the benchmark, we performed some manual analysis of the source conversations in the benchmark data and compared to the gold standard answers. Upon review we noticed several imperfect and inaccurate gold standard answers provided by the benchmark. There are two categories of such ``imperfect" answers: highly inaccurate and partially inaccurate/unclear/incomprehensive. An example of highly inaccurate is the LoCoMo-provided gold standard answer to their question  ``What might John's degree be in?" being ``Political science, Public administration, Public affairs". However, in the same session of where the cited gold standard evidence lies, John shared his certificate of a university degree in Graphic Design. In addition, John mentioned that he lost his job at the mechanical engineering company and might have found a job at a tech company’s hardware team in another dialogue. Coincidentally the answer given by Mnemosyne was ``Mechanical Engineering" for this benchmark question, which was of course marked wrong by the judge LLM despite having a clear argument for being true. As for the question ``What personality traits might Melanie say Caroline has?" the answer is ``Thoughtful, authentic, driven". This is an example of incomprehensive answer since personalities like ``Inspiring, strong, passionate" can also be inferred from conversation sessions other than the ones in the answer field, and these answers are marked wrong by the judge LLM. These indicate that the answers given by LoCoMo can be very inaccurate and the benchmark is inherently flawed. 

We strongly suggest external researchers perform a deeper analysis of the benchmark, such as having a panel of human evaluators directly assess/correct the gold-standard evaluation answers throughout the entire dataset. This also means existing benchmarks will need to be revised. Overall, upon manual review (in the open domain category) of several of Mnemosyne's answers marked incorrect by the judge LLM, we found that Mnemosyne actually provided valid, and in a lot of cases more correct, answers than the gold standard in benchmark. We thus hypothesize that revising the benchmark gold standard answers will not only positively affect Mnemosyne's benchmarking scores but also potentially positively affect other techniques' benchmarking results. This is an important issue to address for both reproducibility and also overall quality of long-term-memory layer systems, since future works from researchers may focus on solving for a flawed benchmark accuracy metric.
\section{Conclusion}
In this paper we presented Mnemosyne, an unsupervised, graph-structured memory system with modular intake filters, probabilistic recall with temporal decay and rewind, and a core summary module that captures persistent user traits. Unlike other techniques, Mnemosyne is specifically designed to run on edge devices, letting agents keep long-horizon memory without relying on brute-force context expansion or heavy cloud infrastructure. We evaluate on the LoCoMo benchmark, for which Mnemosyne beats all methods for temporal reasoning with a score of 60.4\% and overall performs better than almost all leading techniques by scoring an overall J-score of of 54.6\%, coming second only to Memory-R1 despite being a solely edge-based system. Further, while hop retrieval and open-domain are the metrics of focus for several other leading long-term retrieval methods, the focus of our system is on response and interaction naturalness within an atomic domain. This is reflected in our human evaluation, where the full system achieves a 65.8\% win rate over RAG's 31.1\%, consistently producing responses that are more contextually coherent and human like. The ablated Mnemosyne system which omits the core summary system also outperforms RAG in this regard with a win rate of 51.5\%, demonstrating significant efficacy of the traversal and filters specifically. All results were further determined to be statistically significant ($\alpha=0.05$). Future work can revolve around use-based dynamic hyper-parameters. We hope that these results show that highly dynamic heuristics-based memory systems are viable for commercial use, especially given our unsupervised approach which dramatically reduces integration overhead.

\noindent\textbf{Ethics Statement}

All human evaluators gave informed consent to participate in the study, which was conducted under our institutional ethics guidelines. No personally identifiable or health-related information was collected, and all longitudinal healthcare interaction data used to provide scenarios for evaluation was carefully synthesized using a Large Language Model (with real data as reference) as to not correspond with any particular person or patient's journey. The humans surveyed in the human evaluation were all employees of Kaliber AI, raising a potential conflict-of-interest in terms of providing unbiased review metrics. To address this, systems were evaluated in a blind manner with order of system responses further scrambled. A standard set of instructions were given to each subject and no further guidance was given. All completed evaluations were included in the analysis. Further, all other employees and thus humans surveyed were unaware of any details regarding Mnemosyne and the project's purpose even external to the instruction set given, ameliorating any concerns about conflict-of-interest and prior knowledge leakage into results. LLMs were used in a very limited extent to provide help with sentence structure and grammar.

\noindent\textbf{Reproducibility Statement}

The code for Mnemosyne, as well as the engineering scaffolding for execution such as environment setup, Redis database code, and LLM layers, is provided in the supplementary material. Input data and data processing for evaluation, including LoCoMo and retrieval benchmarking, is additionally provided in a separate folder in the code provided in the supplementary materials. Full algorithms are outlined in Appendix \ref{sec: algs}, and our hyperparameter setup for evaluation is provided in Table \ref{tab:hyperparameters}. Mathematical equations for our algorithms and metrics are provided in Appendix \ref{sec: eqs}, and our code implementation will be found under Appendix \ref{sec: downloadable}. We have noticed that reproducibility is a common issue in similar works, so we will prioritize ease and veracity of reproducing our techniques.

\nocite{*}
\bibliographystyle{plainnat}
\bibliography{references}

\newpage
\begin{appendices}

\section{Equations}\label{sec: eqs}

\setcounter{equation}{0}
\renewcommand{\theequation}{A\arabic{equation}}



\textbf{Pruning Score Calculation}. \quad Given node $n$, neighbors $N(n)$, and the weight of edge $e_{nm}$, 
\begin{equation}\label{eq:pruning score}
    \text{PS}(n) = \max_{m\in N(n)}\left(e_{nm}\cdot\tau(e_\text{eff})\right)
\end{equation}

\textbf{Normalized Discounted Cumulative Gain.} \quad nDCG rewards systems that rank relevant memories higher, calculated as
\begin{equation}
    \text{nDCG@}k = \frac{\text{DCG@}k}{\text{IDCG@}k} 
\end{equation}
where $\text{DCG@}k = \sum_{i=1}^{k} \frac{\text{rel}_i}{\log_2(i+1)} $and $\text{rel}_i$ is the relevance score of the retrieved memory at position \textit{i}, and IDCG represents the ideal DCG if all relevant memories were retrieved in perfect order.

\textbf{Win Rate}. \quad Given memory system $S_\text{mem}$,
\begin{equation} \label{eq:win rate}
    \text{WR}(S_\text{mem}) = \frac{\text{number of wins} + 0.5 \cdot \text{number of ties}}{\text{total number of opportunities}}
\end{equation}

\newpage
\section{Algorithms}\label{sec: algs}

\setcounter{algorithm}{0}
\renewcommand{\thealgorithm}{B\arabic{algorithm}}

\begin{algorithm}[H]
\caption{Substance Filter}\label{alg:substance}
\begin{algorithmic}[1]
\Function{HasSubstance}{$s_{new}, G$}
    \State context $\gets s_{new} + G.\text{core\_summary}$
    \State prompt $\gets$ ``Given the context, output 1 if the conversation has substance, else 0."
    \State result $\gets \Call{LLM}{prompt, context}$ \Comment{e.g., Mistral-7B}
    \State \textbf{return} result $= 1$
\EndFunction
\end{algorithmic}
\end{algorithm}

\begin{algorithm}[H]
\caption{Redundancy Filter}\label{alg:redundancy}
\begin{algorithmic}[1]
\Function{DetermineRedundancy}{$n_{new}, G$}
    \State $m^+ \gets \Call{FindMostRedundant}{n_{new}, G}$ 
    \If{$m^+$ is \textbf{null}}
        \State \textbf{return} ADD\_NEW, $n_{new}$
    \EndIf
    \If{$m^+.\text{is\_paired}$ is \textbf{null}}
        \State \textbf{return} PAIR\_NEW, $(n_{new}, m^+)$
    \Else
        \State $n_{j} \gets G.\Call{GetNode}{m^+.\text{is\_paired}}$
        \State $n_{older} \gets \Call{SelectOlder}{m^+, n_{j}}$
        \State $n_{newer} \gets \Call{SelectNewer}{m^+, n_{j}}$
        \State \textbf{return} REPLACE\_AND\_PAIR, $(n_{new}, n_{older}, n_{newer})$
    \EndIf
\EndFunction
\end{algorithmic}
\end{algorithm}

\begin{algorithm}
    \caption{Start Node Selection}\label{alg: start node}
    \begin{algorithmic}[1]
    \Statex \textbf{Input:} Query $q$, Memory graph $G$
    \Statex \textbf{Output:} Start node $n_{best}$
        \Function{SelectStartNode}{$q, G$}
            \State $n_{best} \gets \textbf{null}$
            \State $s_{max} \gets -1$
            \For{each node $n_{i}$ in $G$}
                \State $s_{summary} \gets \Call{CosineSimilarity}{q.\text{embedding}, n_{i}.\text{embedding}}$
                \State $s_{meta} \gets \Call{CosineSimilarity}{q.\text{embedding}, n_{i}.\text{meta\_embedding}}$
                \State $\alpha_{\text{meta}} \gets \Call{CalculateAlpha}{n_{i}.\text{age}}$ \Comment{$\alpha_{\text{meta}}$ increases with age}
                \State $s_{total} \gets \alpha_{\text{meta}} s_{meta} + (1-\alpha_{\text{meta}})s_{summary}$
                \If{$s_{total} > s_{max}$}
                    \State $s_{max} \gets s_{total}$
                    \State $n_{best} \gets n_{i}$
                \EndIf
            \EndFor
            \State \textbf{return} $n_{best}$
        \EndFunction
    \end{algorithmic}
\end{algorithm}



\clearpage
\section{Tables}

\setcounter{table}{0}
\renewcommand{\thetable}{C\arabic{table}}

\begin{table}[ht]
    \centering
    \begin{subtable}[t]{\textwidth}
        \centering
        \begin{tabular}{ |c|c| } 
             \hline
             \textbf{Attribute} & \textbf{Description} \\ 
             \hline
             \texttt{index} & Unique node ID (hash or int) \\ 
             \texttt{neighbors} & List of connected node indices \\ 
             \texttt{is\_paired} & If exists, node ID of paired node \\ 
             \texttt{t\_init} & Time of original conversation \\ 
             \hline
        \end{tabular}
        \caption{Node attributes.}
        \label{tab:graph_node}
    \end{subtable}
    \begin{subtable}[t]{\textwidth}
        \centering
        \begin{tabular}{ |c|c| } 
             \hline
             \textbf{Attribute} & \textbf{Description} \\ 
             \hline
             \texttt{weight} & Similarity score between two nodes \\ 
             \texttt{boost} & Accumulated rewind value \\ 
             \texttt{t\_boosted} & Time of last edge boost during commitment\\ 
             \hline
        \end{tabular}
        \caption{Edge attributes.}
        \label{tab:graph_edge}
    \end{subtable}
    \caption{Attributes of memory graph components.}
    \label{tab:graph_attr}
\end{table}

\begin{table}[ht]
    \centering
    \begin{tabular}{ |c|c| } 
         \hline
         \textbf{Attribute} & \textbf{Description} \\ 
         \hline
         \texttt{summary\_text} & Full summary text \\ 
         \texttt{summary\_vector} & Summary embedding \\ 
         \texttt{hypothetical\_query} & Hypothetical query \\ 
         \texttt{hypothetical\_vector} & Hypothetical query embedding \\ 
         \texttt{keywords} & Keywords \\ 
         \hline
    \end{tabular}
    \medskip
    \caption{Node attributes stored externally.}
    \label{tab:node_attr}
\end{table}

\begin{table}[ht]
\centering
\resizebox{\textwidth}{!}{%
    \begin{tabular}{@{}llcp{0.5\textwidth}@{}}
    \toprule
    \textbf{Component} & \textbf{Parameter} & \textbf{Value} & \textbf{Description} \\
    \midrule
    \multicolumn{4}{l}{\textbf{Graph Construction \& Commitment}} \\
    \midrule
    CommitmentEngine & edge\_threshold & 0.5 & Minimum combined similarity score required to create an edge between two nodes. \\
    CommitmentEngine & $\alpha_{\text{key}}$ & 0.3 & The weight of keyword Jaccard similarity vs. semantic cosine similarity for calculating edge weights. (1-$\alpha_{\text{key}}$)*semantic + $\alpha_{\text{key}}$*keyword. \\
    \midrule
    \multicolumn{4}{l}{\textbf{Redundancy Detection}} \\
    \midrule
    RedundancyDetector & threshold & 0.25 & Score above which a new node is considered redundant with an existing one. \\
    RedundancyDetector & $\alpha_{\text{NMI}}$ & 0.6 & The weight of Normalized Mutual Information (NMI) vs. Jaccard similarity in the base redundancy score. \\
    RedundancyDetector & temporal\_weight\_factor & 0.5 & Multiplier for the temporal bonus, controlling the influence of time proximity on the final redundancy score. \\
    RedundancyDetector & temporal\_decay\_hours & 24 & The ``half-life" (in hours) controlling how quickly the temporal bonus for redundancy fades. \\
    \midrule
    \multicolumn{4}{l}{\textbf{Memory Recall}} \\
    \midrule
    RecallEngine & $\alpha_{\text{meta}}$ & 0.6 & The weight of the semantic (hypothetical query) score vs. the metadata score for selecting the recall start node. \\
    RecallEngine & $\mu$ & 2.0 & Exploration hyperparameter that scales the probability of traversing to a neighbor node during recall propagation. \\
    RecallEngine & mid\_sig & 2,419,200 & The effective age (in seconds, default is 4 weeks) at which the temporal decay function reaches its midpoint. \\
    RecallEngine & floor & 0.05 & The minimum probability (floor) returned by the temporal decay function, ensuring old memories can still be recalled. \\
    RecallEngine & top\_k\_starts & 3 & Number of top nodes to select for multi-start traversal. \\
    RecallEngine & min\_threshold & 0.4 & Minimum score threshold to consider a node relevant. \\
    RecallEngine & max\_nodes\_per\_start & 1 & Maximum nodes to collect from each start point. \\
    \midrule
    \multicolumn{4}{l}{\textbf{Core Summary}} \\
    \midrule
    GetCentralSubset & $\theta_\text{conn}$ & 0.3 & Determines the weight given to a node's centrality, prioritizing memories that are semantically related to many other memories. \\
    GetCentralSubset & $\theta_\text{boost}$ & 0.3 & Governs the influence of memory reinforcement, prioritizing nodes that have been frequently boosted by incoming redundant interactions. \\
    GetCentralSubset & $\theta_\text{rece}$ & 0.2 & Sets the weight for temporal recency, controlling the preference for newer memories during the core summary selection process. \\
    GetCentralSubset & $\theta_\text{ent}$ & 0.2 & Controls the influence of a memory's information density, serving to penalize semantically generic but otherwise well-connected nodes. \\
    GetCentralSubset & $\lambda$ & 28 & Timescale for recency decay (half-life in days) \\
    \bottomrule
    \end{tabular}
}
\medskip
\caption{Hyperparameter settings}
\label{tab:hyperparameters}
\end{table}

\clearpage
\section{Supplementary Evaluations}
\subsection{Retrieval Benchmarking}

\noindent\textbf{Experimental Setup:}
To evaluate Mnemosyne's ability to recall both correct and contextually relevant information, we conducted a retrieval fidelity test between Mnemosyne and a naive RAG baseline, focusing specifically on the start node selection mechanism. Start node selection is the hypothetical bottleneck of our retrieval-based system since all subsequent recall logic is triggered from the start node.

We created a test set of 20 factual, time-specific queries targeting key clinical information across different surgical phases (e.g., ``Where and when was the patient told to complete pre-operative blood tests?"). For each query, we manually defined a list of gold standard memory node IDs representing the ground truth. To ensure a fair comparison, we evaluated only the start node selection mechanism of Mnemosyne against RAG's top-$k$ retrieval, setting $k=3$ for both. Notably, we disabled traversal of Mnemosyne for this benchmark to ensure that we were only measuring the quality of start nodes. This separates the traversal benefit of contextual augmentation using redundancy, boosting, and time-decay from the initial retrieval process.

Two complementary metrics were used to assess performance: (1) hit rate, which measures the percent of queries that resulted in the retrieval of at least one pertinent memory, and (2) normalized discounted cumulative gain (nDCG@3), which accounts for both relevance and ranking position of retrieved memories.

\noindent\textbf{Results:}
Our benchmarking results are shown in Table \ref{tab:benchmarking}. On both metrics, we can observe that Mnemosyne's start node selection mechanism performs better than naive RAG baseline. With a hit rate of 0.75 compared to the naive RAG's 0.64, Mnemosyne showed an absolute increase of 11\% in its ability to correctly retrieve at least one gold standard ID. Additionally, using Mnemosyne increased the nDCG from 0.625 to 0.696, an improvement of 0.071. These improvements indicate that Mnemosyne not only finds relevant memories more accurately, but also positions them more precise in retrieval order. Furthermore, our hybrid scoring approach, which combined semantic similarity with temporal and contextual metadata was validated. The improvement over pure semantic similarity search (RAG) demonstrates that incorporating temporal and contextual information into retrieval process results in better memory selection, even before applying graph traversal.
\begin{table}[ht]
\caption{Retrieval benchmarking results. The strongest results are bolded.}
    \label{tab:benchmarking}
    \centering
    \begin{tabular}{c|c|c}
         System & Hit Rate & nDCG \\
         \hline
         \textbf{Mnemosyne} & \textbf{0.75} & \textbf{0.696} \\
         Naive RAG & 0.64 & 0.625 \\
    \end{tabular}
\end{table}

\noindent\textbf{Commentary:}
This evaluation starts with framing start-node selection as the main bottleneck of the system and compares it directly to standard RAG. Even in this stripped down form, Mnemosyne slightly beats RAG on hit rate and nDCG which acts as a good sanity check that the base retrieval mechanism is sound. The rest of the system, including traversal, filters, and core summary, act as augmentations that drive improvements in naturalness as the main body analysis shows.

\clearpage
\section{Downloadable Supplementary Material (Code)} \label{sec: downloadable}
Code release is pending legal approval but will be updated here.

\end{appendices}

\end{document}